\documentclass{amsproc}
\usepackage[utf8]{inputenc}
\usepackage{tikz}
\usepackage{geometry}
\usepackage{hyperref}

\title{The effects of data size on Automated Essay Scoring engines}
\author{Christopher Ormerod}
\email{christopher.ormerod@cambiumassessment.com}
\author{Amir Jafari}
\email{amir.jafari@cambiumassessment.com}
\author{Susan Lottridge}
\email{susan.lottridge@cambiumassessment.com}
\author{Milan Patel}
\email{milan.patel@cambiumassessment.com}
\author{Amy Harris}
\email{aharris.air@gmail.com}
\author{Paul van Wamelen}
\email{paul.vanwamelen@cambiumassessment.com}
\date{October 2020}

\begin{document}

\maketitle

\begin{abstract}
We study the effects of data size and quality on the performance on Automated Essay Scoring (AES) engines that are designed in accordance with three different paradigms; A frequency and hand-crafted feature-based model, a recurrent neural network model, and a pretrained transformer-based language model that is fine-tuned for classification. We expect that each type of model benefits from the size and the quality of the training data in very different ways. Standard practices for developing training data for AES engines were established with feature-based methods in mind, however, since neural networks are increasingly being considered in a production setting, this work seeks to inform us as to how to establish better training data for neural networks that will be used in production. 
\end{abstract}

\section{Introduction}


Constructing an essay engages a student's ability to critically think, analyze, organize, and synthesize ideas. This means that the assessment of student essays is a valuable way to test the upper levels of Bloom's taxonomy \cite{blooms}. As such, essay items are often important tools in any comprehensive assessment program. Having essays scored professionally in a standardized testing program comes at a great cost to the states \cite{StandardTestingCost}. Factors that influence how the states choose to assess students include increased testing and a general lack of state funds. An Automated Essay Scoring (AES) engine is a statistical model used to evaluate an essay in a manner that is as close to human scoring as possible. The cost of using AES engines has been estimated to be from one fifth to a half of the cost of human scoring \cite{ASG}. 

To the best of our knowledge, the majority of AES engines currently in production use linguistic features with or without Latent Semantic Analysis (LSA) \cite{erater2, LSA, TCT, PEG}. We generally consider these to be in the class of Bag-of-Words (BOW) methods. The performance of these engines is focused on the selection and design of useful features. These features can require a great deal of work to test, extrapolate, and implement. Once a suitable set of features is chosen, a classical machine-learning classifier is fit to a set of training data to obtain an AES engine. Features and knowledge of the classifier weights can be beneficial to the generation of feedback for teachers and students, but also in diagnosing why an engine may have given the score it has assigned. The downside of such engines is that they tend to be brittle because language is not adequately encompassed by a finite collection of linguistic features. These engines are modelled on vocabulary observed in the training sample and upon a small collection of weights. Increasing the amount of training data should increase this vocabulary and provide more accurate estimates for an optimal set of weights. 

Technology that has seen far less use in production has been neural networks \cite{AttAES, OurBERT, NNAES}. The first neural network-based AES engines to appear in research were based on a mix of convolutional and recurrent layers with attention \cite{AttAES, NNAES}. These engines used word-embeddings which map words to a semantic vector space \cite{fasttext}. Unlike in LSA, word order is vitally important. These models involve millions of parameters and learn linguistic features implicitly rather than explicitly. These models generally require a lot of data to train, however, we expect that these methods can be more accurate with sufficient data. 

The field of neural networks applied Natural Language Processing (NLP) has undergone a revolution since the development of the transformer \cite{AttentionNeed}. Where LSTM-based models require a lot of data to train, pretrained transformer-based language models can take advantage of vast corpora of unsupervised data. As such, transformer-based models often involve an order of magnitude more parameters that store pre-baked features. We think of transformer-based models as having a general understanding of language before being fine-tuned for a classification task. This pretraining and the number of parameters involved should affect the size of data required to make inferences in an AES setting. As a benchmark, we consider AES engines based on the BERT architecture \cite{BERTAES, OurBERT}.

We generally possess two types of labels for essay data depending on how the essays were assessed; single-scored data, where the data has been scored once for assessment purposes by teachers or assessment companies, and double scored data, which has been read by two independent professionally trained assessors \cite{Framework}. This gives us labelled data of two distinct levels of quality. There is an assumption that the high-quality data leads to a high-quality engine, however, this assumption and the framework for implementing AES systems was written with feature-based methods in mind. There are valid reasons that this assumption may break down for neural networks. The neural networks in question are capable of modelling an entire essay prompt. The more language a neural network is exposed to, the greater the number of patterns/features it learns. Testing agencies generally possess far more single-scored than double-scored data, hence, it is our hypothesis that the vast quantities of single-scored data may add value in the context of neural network-based engines. We show that we may use single-scored data to enhance the results of engines trained on double-scored data.

The use of AES is not without controversy with many citing the ability to game the system in certain ways that are not necessarily conducive to good writing \cite{babel, Les}. The common critiques of AES are also predominately directed toward feature-based techniques \cite{Les}. We foresee that due to the techniques in NLP becoming increasingly sophisticated and accurate that AES engines will become more robust to being gamed in this way. Furthermore, recent advances in language models show the potential to decrease costs while simultaneously increasing the overall quality of scoring. These approaches can also be trained to be more robust, but still susceptible to gaming \cite{NNgaming}. The biggest downside of neural network-based models that the features are buried in a sea of parameters and are not as transparent as feature-based models. 

In \S \ref{sec:experiment} we discuss the various experiments. In \S \ref{sec:results} we will review the results of the experiments. Finally in \S \ref{sec:discussion} we will discuss some broader consequences, including limitations of the approach we took and the broader landscape of the applications of NLP in an assessment setting.

\section{Experimental Design}\label{sec:experiment}

An essay is typically assigned a score between 0 and 10 reflecting how well the essay is written. A good essay should be organized logically, flow smoothly, and explain a central idea. It is also important to correctly spell words, adhere to the rules of grammar, and appropriately punctuate sentences. These desired qualities have guided the development of a rubric that assesses essays with respect to three different traits \cite{Rubric}:
\begin{itemize}
    \item Organization/Purpose: This measures how well-focused an essay is, how well the author uses citations and transitions, and how well the introduction and conclusion fit with the essay. This trait is scored out of 4.
    \item Elaboration: This measures clarity/readability, how engaging the essay is, and an appropriate vocabulary. This trait is scored out of 4.
    \item Conventions: This measures the correctness of spelling and grammar. This trait is scored out of 2. 
\end{itemize}
The final score out of 10 is the sum of the individual scores for each trait. When using AES, for each essay item, we typically design three separate engines which are fit to the traits rather than the final score. Due to the nature of the traits, we expect that the three different types of engines (BOW, LSTM, BERT) benefit from more data differently for each trait. In particular, we expect that the BOW models perform well even with a small amount of data, however, neural network-based models should provide better performance once they are provided sufficient data. Furthermore, the appropriate parameters governing the implicit features of a BERT model will require more data to become effective, and we expect that the number of effective features an LSTM will learn from a small dataset will not be large enough to effectively model any of the traits.

In developing an engine for a particular item/essay prompt, the first step is to obtain a sufficiently large corpus of text responses to be used as a training set. Each text is assessed independently for each of the three traits by two human raters. If both the human raters agree, their agreed upon score is reported, while if the two raters disagree, the score is adjudicated by a third reader, and sometimes even a fourth if the third reader disagrees with both scores given depending on the rules set out by the agency governing the scoring \cite{Eval}. The aim of having two raters with adjudication and resolution is that the final score is as close as feasible to being a true interpretation of the essay rubric for each trait. While it takes more than twice the time to score in this way, the argument is that the better the labels are, the better the AES engine is. 

In addition to obtaining more accurate scores, two human raters allow us to obtain a measure of how well the item was scored from a psychometric standpoint by gauging inter-rater reliability statistics \cite{IRR}. The most important metric to evaluate is the Cohen's quadratic weighted kappa (QWK) statistic, defined by
\begin{equation}\label{eq:qwk}
\kappa = \frac{\sum \sum w_{ij} x_{ij}}{\sum \sum w_{ij} m_{ij}}
\end{equation}
where $x_{i,j}$ is the observed probability
\[
m_{i,j} = x_{ij}(1-x_{ij}),
\]
and
\[
w_{ij} = 1- \frac{(i-j)^2}{(k-1)^2},
\]
where $k$ is the number of classes. 

There are several reasons QWK is favored over accuracy. The first reason is that QWK depends on the entire confusion matrix, not just the diagonal elements. This means that larger score discrepancies have a greater effect on the QWK than smaller ones, whereas accuracy considers all incorrect scores equally. Secondly, the use of observed probabilities in (\ref{eq:qwk}) have the effect of taking into account the rarity of the score. One interpretation of this metrics is that the QWK captures the level of agreement above and beyond what would be obtained by chance and weighted by the extent of disagreement. 

The QWK and accuracy are not the only two metrics that are used in the calibration of an AES engine; we also consider the standardized mean difference (SMD) which measures the overall spread of scores. The SMD is defined by
\begin{equation}
SMD(s_1, s_2) = \frac{|\mu(s_1) - \mu(s_2) |}{\sqrt{\frac{\sigma(s_1)^2}{2} + \frac{\sigma(s_2)^2}{2} }},
\end{equation}
where $\mu$ and $\sigma$ are the mean and standard deviation functions. 

The framework outlined in \cite{Framework} recommends that the QWK between two raters should be above $0.7$ before being considered as a training set for an AES engine. We generally require that the difference between the QWK between two human raters and the QWK between the AES engine and the final human resolved score should be no greater than $0.1$. Furthermore, the SMD between the raters and the SMD between the AES engine and the final score should be less than $0.15$. It is for this reason that we need to consider both the SMD and the QWK as we increase the amount of training data.

In our first experiment, we seek to determine how our models improve with the amount of data provided. For this experiment, we will use a large corpus of single-scored data and gradually increase the amount of single scored data used for training. We consider the following two sets of data:
\begin{itemize}
    \item Training: We use a corpus of 15,000 single-scored responses, each of which has been assessed in each of the three traits for assessment purposes. 
    \item Validation: We use an additional 2,000 responses from the same source as held-out single-scored validation data. 
\end{itemize}
The training data was divided into a chain of 30 subsets, $\{ X_i : 1 \leq i \leq 30\}$ so that $X_i \subset X_{i+1}$ and $|X_{i+1}| - |X_i| = 500$. That is to say we have a chain 30 subsets whose sizes range between 500 and 15,000 in steps of 500. To determine how well the average model does in comparison with humans, we use 5-fold validation by further subdividing each subset into 5 different test/train splits. Each of subset, $X_i$, is the disjoint union of a test/train-split and each subset, $X_i$, is the disjoint union of the 5 different test sets. Our final QWK is the average over the folds. In this way we determine how each of the types of models responds to increases in data size.


We consider the final performance to be the average of the QWK on the held-out set for each of the folds \cite{NNAES}. It is important to note that we are not gauging the best possible performance, which is the goal of most research programs in neural networks. This study seeks to gauge average performance, so averaging over the folds has the effect of smoothing out the variability in the resulting QWK. We often found, in the case of the LSTM, that the engine failed to converge altogether, which are results we have not discarded for the fidelity of the experiment. One of the other factors that influences the variability of the QWK measurement is the rarity of scores. Failing to predict rare scores has a more dramatic effect on the QWK than failing to predict common scores. For this reason, we present the score distributions for the validation set in Table \ref{tab:score_distributions}.

\begin{table}[!ht]
    \centering
    \begin{tabular}{| c | c | c || c | c | c | c | c |}\hline
        &&& \multicolumn{5}{c|}{Score} \\ \hline
        Prompt & Validation & Dimension & 0 & 1 & 2 & 3 & 4 \\ \hline\hline
                Essay \#1 & Single-Scored & Elaboration & 10.30\% & 35.6\% & 34.55\% & 18.85\% & 0.7\% \\
         &  & Organization & 7.20\% & 32.1\% & 44.95\% & 15.2\% & 0.55\% \\
          &  & Conventions & 8.80\% & 20.05\% & 71.15\% && \\\hline
        Essay \#2 & Single-Scored & Elaboration & 12.8 \% & 19.3\% & 49.4\% & 14.75\% & 3.75\% \\
         &  & Organization & 8.2\% & 16.4\% & 46.45\% & 23.6\% & 5.35\% \\
          &  & Conventions & 5.3\% & 20.75\% & 73.95\% && \\\hline
    \end{tabular}
    \caption{The score distribution for the two validation sets for essay prompts \#1 and \# 2.}
    \label{tab:score_distributions}
\end{table}

In counting the number of models in this experiment, we note that we have a different model for each dimension, subset, and fold. In total, we require an evaluation of 450 models for each type of model and for each essay item, making 2700 models in total (or 900 for each type of model used). Due to the sheer number of models involved, it was not feasible to perform hyper-parameter tuning on this scale. For this reason the results we present are not the best possible results with each architecture. They are to be considered a reflection of how each architecture scales as the result of a single model resulting from generically chosen parameters. The evaluation of these 2700 models on the validation set should give us a clear idea of how the size of data affects the quality of the engine.

Our second experiment challenges the long-standing assumption that it is more important to have small amount of good quality data than it is to have a large quantity of poorly labelled data. To test this, we took the same two essay prompts from the first experiment in which we had a large corpus (approximately 50k) of both single-scored data and approximately 2500 double scored data that was designed to be used to build an AES engine. This gives us two distinct qualities for the labels used in training. In this experiment, we have three sets of data:
\begin{itemize}
    \item Single-scored Training: We use approximately the full set of 50k responses as training data with their corresponding labels. 
    \item Double-scored Training: We use approximately 2000 responses with their final resolved score as the labels used in training. 
    \item Validation: We use the remaining 500 double scored data, with their final resolved score as the labels to validate machines on both sets of training data.
\end{itemize}
The inter-rater reliability statistics for the validation set for the two human reads are presented in table \ref{tab:irr2}.

\begin{table}[!ht]
    \centering
    \begin{tabular}{|c|c |c | c | c | c | c |} \hline
    \multicolumn{4}{|c|}{} &\multicolumn{3}{c|}{H1-H2} \\ \hline
        &Count & dimension & max score & QWK & Acc & SMD \\ \hline 
    \# 1 &479& Organization &4& 0.828 & 74.2\% & 0.009 \\ 
         & & Elaboration &4 &0.760 & 70.1\% & 0.044 \\ 
         & & Conventions &2& 0.745 & 77.2\% & 0.030\\ \hline
    \# 2 &481 & Organization &4& 0.840 & 78.4\% & 0.036\\ 
         & & Elaboration &4& 0.825 & 77.3\% & 0.010 \\ 
         & & Conventions &2& 0.756 & 79.0\% & 0.054\\ \hline
    \end{tabular}
    \caption{This table summarizes the inter-rater reliability statistics for the double-scored data for the two essay items chosen for this study. The H1-H2 refers to human raters 1 and 2.}
    \label{tab:irr2}
\end{table}

In this experiment, we have two sets of models, one trained on the single-scored data and another trained on the double-scored data. Our hypothesis is that training on 2000 double-scored data performs better than a large quantity of single-scored data. There are good reasons this hypothesis could be false, especially for neural networks. 

There is one issue in this comparison which need to be addressed regarding the nature of the data; are the single-scored labels/scores an accurate representation of the double-scored labels/scores? We know that the corpus of single-scored responses originates from a time in which all responses were scored by a human. While we also know that the double-scored responses were drawn from the same sample, the administrative conditions for assessing responses may have changed between the time the corpus was originally assessed and the time the data was assessed for the purposes of building an AES engine. The average scores and spread of scores could differ which would adversely affect the SMD and the QWK to a lesser extent. What should be true is that the single-scored data should be able to form a first approximation for the double-scored data. For this reason, we consider an extra step; we use the models obtained by training on the single-scored data to define a set of initial weights to be used for training on the double-scored data. In this way, we test the overall usefulness of the single-scored data in training for a AES for use in production. To our knowledge, this type of data is completely disregarded in the development of AES engines.  

We now describe the models used in the specific AES engines to test our hypotheses:
\begin{itemize}
\item{BOW: The BOW-based engine may be considered an ensemble between an LSA-based engine and a feature-based engine. In this engine we need to choose the linguistic features to include and the LSA dimension. We include a list of sixteen features such as the number of punctuation errors, misspellings, typos and average sentence length. Since we expect that the conventions score is mainly dependent on these features, we chose an LSA dimension of 10 for conventions, while we expect that the elaboration and organization dimensions are driven by the semantic content, so we chose an LSA dimension of 70 for these dimensions. The resulting twenty six or ninety six features are then concatenated and an ordinal probit model is applied to produce a classification.}
\item{LSTM: To evaluate an LSTM-based architecture we first established an embedding. We took a large corpus of student texts, tokenized them with respect to the standard spaCy tokenizer and formed case insensitive fastText embedding \cite{fasttext}. This embedding was used to transform the inputs into a two layer bidirectional LSTM with 400 hidden units in each direction in each layer. A simplified attention mechanism consisting of a weighted average of the output of the LSTM was applied with a linear layer to form the output. Optimization with the adaptive minimization algorithm, Adam, with standard learning rates. These models were implemented using Pytorch. No pretraining was involved.}
\item{BERT: To evaluate pretrained transformer-based architectures we chose the standard BERT architecture. For conventions, the cased version of the base architecture with 12 layers while both the elaboration and organization scores used the uncased version of the same architecture. These models were obtained and fine-tuned using the codebase of Hugginface\footnote{https://github.com/huggingface/transformers}. A version of Adam was used with a standard learning rate.}
\end{itemize}
 
\section{Results}\label{sec:results}

\begin{figure}[!ht]
\centering
Elaboration

\begin{tikzpicture}[scale=0.5]
\node at (5,-2) {Essay \# 1};
\draw[thick,->] (0,0) -- (10,0);
\draw[thick,->] (0,0) -- (0,10);
\draw[black!40] (0,0) grid (10,10);
\node at (0,-1) {0};
\node at (-1,0) {0.3};
\node at (2,-1) {3k};
\node at (-1,2) {0.4};
\node at (4,-1) {6k};
\node at (-1,4) {0.5};
\node at (6,-1) {9k};
\node at (-1,6) {0.6};
\node at (8,-1) {12k};
\node at (-1,8) {0.7};
\node at (10,-1) {15k};
\node at (-1,10) {0.8};
\draw[thick, red] (0.333,0.000)  -- (0.667,0.000) -- (1.000,1.438) -- (1.333,2.245) -- (1.667,4.890) -- (2.000,5.374) -- (2.333,5.738) -- (2.667,5.850) -- (3.000,5.877) -- (3.333,6.140) -- (3.667,6.062) -- (4.000,6.094) -- (4.333,6.291) -- (4.667,6.059) -- (5.000,6.232) -- (5.333,6.277) -- (5.667,6.341) -- (6.000,6.473) -- (6.333,6.515) -- (6.667,6.595) -- (7.000,6.730) -- (7.333,6.679) -- (7.667,6.863) -- (8.000,6.741) -- (8.333,6.891) -- (8.667,6.828) -- (9.000,6.853) -- (9.333,6.916) -- (9.667,6.993) -- (10.000,7.109);
\draw[thick, blue] (0.333,6.303)  -- (0.667,6.161) -- (1.000,6.277) -- (1.333,6.363) -- (1.667,6.452) -- (2.000,6.554) -- (2.333,6.450) -- (2.667,6.526) -- (3.000,6.552) -- (3.333,6.612) -- (3.667,6.585) -- (4.000,6.551) -- (4.333,6.537) -- (4.667,6.609) -- (5.000,6.614) -- (5.333,6.599) -- (5.667,6.595) -- (6.000,6.582) -- (6.333,6.579) -- (6.667,6.627) -- (7.000,6.638) -- (7.333,6.621) -- (7.667,6.648) -- (8.000,6.641) -- (8.333,6.631) -- (8.667,6.631) -- (9.000,6.625) -- (9.333,6.628) -- (9.667,6.623) -- (10.000,6.683);
\draw[thick, black] (0.333,5.187)  -- (0.667,6.145) -- (1.000,6.218) -- (1.333,6.637) -- (1.667,6.851) -- (2.000,6.488) -- (2.333,6.721) -- (2.667,6.726) -- (3.000,6.585) -- (3.333,7.033) -- (3.667,7.100) -- (4.000,7.254) -- (4.333,7.160) -- (4.667,7.068) -- (5.000,7.216) -- (5.333,7.084) -- (5.667,7.182) -- (6.000,7.136) -- (6.333,7.135) -- (6.667,7.234) -- (7.000,7.209) -- (7.333,7.144) -- (7.667,7.304) -- (8.000,7.107) -- (8.333,7.048) -- (8.667,7.047) -- (9.000,7.054) -- (9.333,7.087) -- (9.667,7.171) -- (10.000,7.171);
\end{tikzpicture}
\begin{tikzpicture}[scale=0.5]
\node at (5,-2) {Essay \# 2};
\draw[thick,->] (0,0) -- (10,0);
\draw[thick,->] (0,0) -- (0,10);
\draw[black!40] (0,0) grid (10,10);
\node at (0,-1) {0};
\node at (-1,0) {0.3};
\node at (2,-1) {3k};
\node at (-1,2) {0.4};
\node at (4,-1) {6k};
\node at (-1,4) {0.5};
\node at (6,-1) {9k};
\node at (-1,6) {0.6};
\node at (8,-1) {12k};
\node at (-1,8) {0.7};
\node at (10,-1) {15k};
\node at (-1,10) {0.8};
\draw[thick, blue] (0.333,6.119)  -- (0.667,6.342) -- (1.000,6.689) -- (1.333,6.853) -- (1.667,6.917) -- (2.000,7.127) -- (2.333,6.990) -- (2.667,6.994) -- (3.000,6.947) -- (3.333,6.978) -- (3.667,6.961) -- (4.000,6.995) -- (4.333,7.093) -- (4.667,6.985) -- (5.000,7.018) -- (5.333,7.033) -- (5.667,7.064) -- (6.000,7.026) -- (6.333,7.013) -- (6.667,7.113) -- (7.000,7.141) -- (7.333,7.110) -- (7.667,7.125) -- (8.000,7.191) -- (8.333,7.178) -- (8.667,7.215) -- (9.000,7.272) -- (9.333,7.281) -- (9.667,7.279) -- (10.000,7.348);
\draw[thick, red] (0.333,0.000)  -- (0.667,0.000) -- (1.000,0.000) -- (1.333,0.000) -- (1.667,0.000) -- (2.000,0.000) -- (2.333,0.282) -- (2.667,1.110) -- (3.000,2.560) -- (3.333,5.065) -- (3.667,5.331) -- (4.000,5.622) -- (4.333,5.718) -- (4.667,5.668) -- (5.000,5.895) -- (5.333,5.687) -- (5.667,6.015) -- (6.000,6.335) -- (6.333,6.063) -- (6.667,6.141) -- (7.000,6.437) -- (7.333,6.426) -- (7.667,6.783) -- (8.000,6.760) -- (8.333,6.892) -- (8.667,7.156) -- (9.000,7.062) -- (9.333,7.015) -- (9.667,7.032) -- (10.000,7.428);
\draw[thick, black] (0.333,5.529)  -- (0.667,6.862) -- (1.000,7.116) -- (1.333,7.348) -- (1.667,7.857) -- (2.000,7.919) -- (2.333,7.907) -- (2.667,7.890) -- (3.000,7.759) -- (3.333,7.989) -- (3.667,7.965) -- (4.000,8.024) -- (4.333,7.878) -- (4.667,7.843) -- (5.000,8.092) -- (5.333,8.018) -- (5.667,8.021) -- (6.000,8.435) -- (6.333,8.196) -- (6.667,8.350) -- (7.000,8.319) -- (7.333,8.427) -- (7.667,8.530) -- (8.000,8.294) -- (8.333,8.752) -- (8.667,8.494) -- (9.000,8.524) -- (9.333,8.684) -- (9.667,8.667) -- (10.000,8.7);
\end{tikzpicture}

\begin{tikzpicture}
\draw[thick, blue] (0,0) --(.5,0);
\node at (1.2,0) {BOW};
\end{tikzpicture}\hspace{1cm}
\begin{tikzpicture}
\draw[thick, red] (0,0) --(.5,0);
\node at (1.2,0) {LSTM};
\end{tikzpicture}\hspace{1cm}
\begin{tikzpicture}
\draw[thick, black] (0,0) --(.5,0);
\node at (1.2,0) {BERT};
\end{tikzpicture}
\caption{The performance on the dimension of elaboration. On the left we have the essay item \# 1, on the right we have essay item \#2.}\label{fig:val_elaboration}
\end{figure}
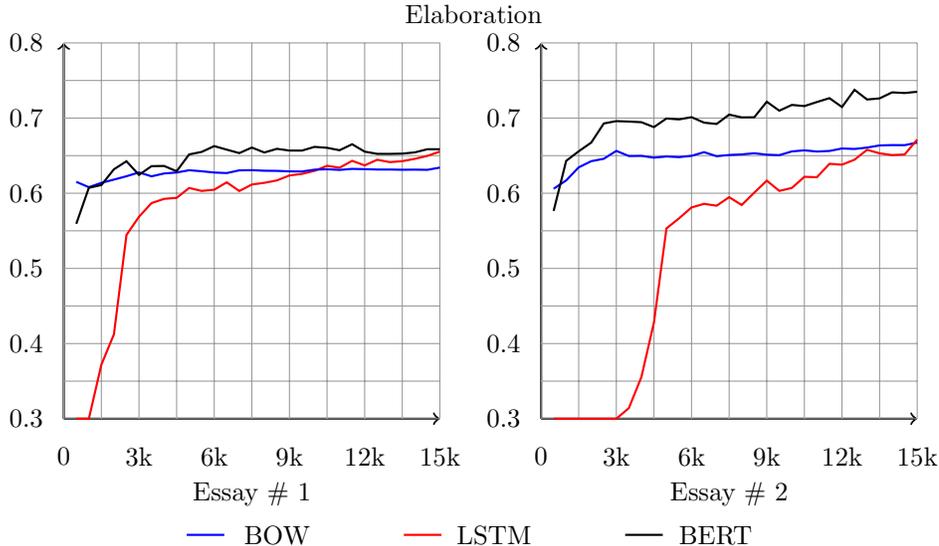

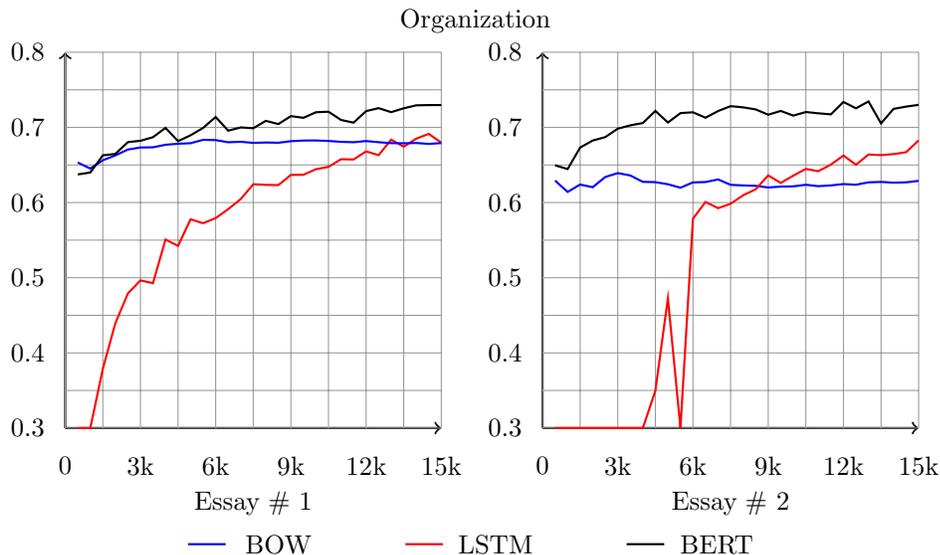
\begin{figure}[!ht]
\centering
Organization

\begin{tikzpicture}[scale=0.5]
\node at (5,-2) {Essay \# 1};
\draw[thick,->] (0,0) -- (10,0);
\draw[thick,->] (0,0) -- (0,10);
\draw[black!40] (0,0) grid (10,10);
\node at (0,-1) {0};
\node at (-1,0) {0.3};
\node at (2,-1) {3k};
\node at (-1,2) {0.4};
\node at (4,-1) {6k};
\node at (-1,4) {0.5};
\node at (6,-1) {9k};
\node at (-1,6) {0.6};
\node at (8,-1) {12k};
\node at (-1,8) {0.7};
\node at (10,-1) {15k};
\node at (-1,10) {0.8};
\draw[thick,red] (0.333,0.000)  -- (0.667,0.000) -- (1.000,1.574) -- (1.333,2.789) -- (1.667,3.582) -- (2.000,3.929) -- (2.333,3.855) -- (2.667,5.016) -- (3.000,4.847) -- (3.333,5.555) -- (3.667,5.450) -- (4.000,5.586) -- (4.333,5.824) -- (4.667,6.093) -- (5.000,6.487) -- (5.333,6.471) -- (5.667,6.464) -- (6.000,6.735) -- (6.333,6.739) -- (6.667,6.887) -- (7.000,6.950) -- (7.333,7.153) -- (7.667,7.146) -- (8.000,7.361) -- (8.333,7.261) -- (8.667,7.673) -- (9.000,7.486) -- (9.333,7.700) -- (9.667,7.828) -- (10.000,7.593);
\draw[thick,blue] (0.333,7.065)  -- (0.667,6.904) -- (1.000,7.125) -- (1.333,7.254) -- (1.667,7.411) -- (2.000,7.465) -- (2.333,7.470) -- (2.667,7.537) -- (3.000,7.565) -- (3.333,7.579) -- (3.667,7.669) -- (4.000,7.662) -- (4.333,7.604) -- (4.667,7.617) -- (5.000,7.588) -- (5.333,7.598) -- (5.667,7.591) -- (6.000,7.633) -- (6.333,7.650) -- (6.667,7.652) -- (7.000,7.639) -- (7.333,7.616) -- (7.667,7.605) -- (8.000,7.637) -- (8.333,7.606) -- (8.667,7.581) -- (9.000,7.578) -- (9.333,7.585) -- (9.667,7.561) -- (10.000,7.583);
\draw[thick,black] (0.333,6.749)  -- (0.667,6.799) -- (1.000,7.256) -- (1.333,7.294) -- (1.667,7.608) -- (2.000,7.644) -- (2.333,7.739) -- (2.667,7.987) -- (3.000,7.634) -- (3.333,7.790) -- (3.667,7.986) -- (4.000,8.277) -- (4.333,7.913) -- (4.667,7.999) -- (5.000,7.978) -- (5.333,8.174) -- (5.667,8.089) -- (6.000,8.300) -- (6.333,8.255) -- (6.667,8.403) -- (7.000,8.417) -- (7.333,8.198) -- (7.667,8.126) -- (8.000,8.433) -- (8.333,8.512) -- (8.667,8.406) -- (9.000,8.508) -- (9.333,8.587) -- (9.667,8.594) -- (10.000,8.594);
\end{tikzpicture}
\begin{tikzpicture}[scale=0.5]
\node at (5,-2) {Essay \# 2};
\draw[thick,->] (0,0) -- (10,0);
\draw[thick,->] (0,0) -- (0,10);
\draw[black!40] (0,0) grid (10,10);
\node at (0,-1) {0};
\node at (-1,0) {0.3};
\node at (2,-1) {3k};
\node at (-1,2) {0.4};
\node at (4,-1) {6k};
\node at (-1,4) {0.5};
\node at (6,-1) {9k};
\node at (-1,6) {0.6};
\node at (8,-1) {12k};
\node at (-1,8) {0.7};
\node at (10,-1) {15k};
\node at (-1,10) {0.8};
\draw[thick,blue] (0.333,6.584)  -- (0.667,6.283) -- (1.000,6.479) -- (1.333,6.408) -- (1.667,6.680) -- (2.000,6.784) -- (2.333,6.721) -- (2.667,6.551) -- (3.000,6.543) -- (3.333,6.487) -- (3.667,6.395) -- (4.000,6.536) -- (4.333,6.546) -- (4.667,6.615) -- (5.000,6.472) -- (5.333,6.455) -- (5.667,6.445) -- (6.000,6.401) -- (6.333,6.424) -- (6.667,6.431) -- (7.000,6.473) -- (7.333,6.436) -- (7.667,6.456) -- (8.000,6.492) -- (8.333,6.474) -- (8.667,6.538) -- (9.000,6.549) -- (9.333,6.529) -- (9.667,6.538) -- (10.000,6.577);
\draw[thick,red] (0.333,0.000)  -- (0.667,0.000) -- (1.000,0.000) -- (1.333,0.000) -- (1.667,0.000) -- (2.000,0.000) -- (2.333,0.000) -- (2.667,0.000) -- (3.000,0.993) -- (3.333,3.456) -- (3.667,0.000) -- (4.000,5.572) -- (4.333,6.016) -- (4.667,5.851) -- (5.000,5.967) -- (5.333,6.194) -- (5.667,6.353) -- (6.000,6.723) -- (6.333,6.522) -- (6.667,6.716) -- (7.000,6.895) -- (7.333,6.834) -- (7.667,7.010) -- (8.000,7.252) -- (8.333,7.005) -- (8.667,7.279) -- (9.000,7.263) -- (9.333,7.289) -- (9.667,7.339) -- (10.000,7.655);
\draw[thick,black] (0.333,6.992)  -- (0.667,6.891) -- (1.000,7.464) -- (1.333,7.650) -- (1.667,7.741) -- (2.000,7.966) -- (2.333,8.054) -- (2.667,8.116) -- (3.000,8.439) -- (3.333,8.130) -- (3.667,8.380) -- (4.000,8.402) -- (4.333,8.257) -- (4.667,8.435) -- (5.000,8.565) -- (5.333,8.533) -- (5.667,8.479) -- (6.000,8.337) -- (6.333,8.438) -- (6.667,8.314) -- (7.000,8.409) -- (7.333,8.375) -- (7.667,8.347) -- (8.000,8.677) -- (8.333,8.508) -- (8.667,8.690) -- (9.000,8.107) -- (9.333,8.493) -- (9.667,8.553) -- (10.000,8.6);
\end{tikzpicture}

\begin{tikzpicture}
\draw[thick, blue] (0,0) --(.5,0);
\node at (1.2,0) {BOW};
\end{tikzpicture}\hspace{1cm}
\begin{tikzpicture}
\draw[thick, red] (0,0) --(.5,0);
\node at (1.2,0) {LSTM};
\end{tikzpicture}\hspace{1cm}
\begin{tikzpicture}
\draw[thick, black] (0,0) --(.5,0);
\node at (1.2,0) {BERT};
\end{tikzpicture}
\caption{The performance on the organization trait. We have set a lower bound of 0.4 and an upper bound of 0.8 to better display the differences in the models.}\label{fig:val_organization}
\end{figure}

In our first experiment, we are interested in how these engines perform as we increase the amount of data, we provide for training from 500 responses up to 15k responses in increments of 500. We compare the three different engine types on two separate essay prompts, each of which has three traits. In our view, elaboration and organization are grouped together, as they depend on similarly defined and intersecting features, while conventions are defined on an almost disjoint set of features. We start by examining the performance on Elaboration and Organization; the change in the resulting QWK on the validation data for the traits of Elaboration and Organization is presented in Figure \ref{fig:val_elaboration} and Figure \ref{fig:val_organization} respectively.

One of the difficulties that AES engines have in assessing Elaboration and Organization is that these two traits should be scored independently of the spelling and grammar. This poses a unique difficulty as these two traits benefit from how the AES engines may extrapolate the correct word when one is incorrectly spelled. To adjust for this, the LSA component of the BOW engine was subjected to spell-correction while spell-correction was not applied to the features component of the model.

For both prompts, the performance of the LSTM-based AES engine is heavily dependent on the amount of data. It is clear that when the LSTM starts with little to no data, the models do poorly, however, it is clear that with enough data, the LSTM-based engines can perform comparably with the BOW-based models and in most cases, these engines exceed the performance of BOW-based models. One of the distinct advantages the LSTM model possesses is the use of the fast-text embedding. Since fast-text embeddings are the result of an average over subwords, if sufficiently many subwords in an incorrectly spelled word are present, it is possible that the averaging mechanism may approximate the meaning of the misspelled word \cite{fasttext}. One of the problems in this model did not always converge, especially in the case when we used smaller datasets. Most of the time these models would have been discarded in the hyperparameter selection, however, since we did not tune the LSTM models, we see a considerable drop in performance on Essay \#2 due to this instability. It was clear that the larger the dataset, the more stable the results were.

As expected, the BOW-based models show much less of a performance increase as we increase the quantity of data. The size and accuracy of the LSA component should account for some increase in accuracy, however, in some cases, there is an even a slight drop in performance of the BOW-based model as we increase the amount of data used. 

In most cases, the BERT engine not only improves with the increase in data, but it also shows very solid performance with very little data. This indicates that the pretraining endows the BERT engine with sufficiently many features to be useful for conventions before training begins. These results seem to be different from those in a previous study, but, the nature of the data is very different \cite{SmallDataBERT}. 

\begin{figure}[!ht]
    \centering

Conventions

\begin{tikzpicture}[scale=0.5]
\node at (5,-2) {Essay \#1};
\draw[thick,->] (0,0) -- (10,0);
\draw[thick,->] (0,0) -- (0,10);
\draw[black!40] (0,0) grid (10,10);
\node at (0,-1) {0};
\node at (-1,0) {0.3};
\node at (2,-1) {3k};
\node at (-1,2) {0.4};
\node at (4,-1) {6k};
\node at (-1,4) {0.5};
\node at (6,-1) {9k};
\node at (-1,6) {0.6};
\node at (8,-1) {12k};
\node at (-1,8) {0.7};
\node at (10,-1) {15k};
\node at (-1,10) {0.8};
\draw[thick,red] (0.333,0.000)  -- (0.667,0.000) -- (1.000,0.000) -- (1.333,0.000) -- (1.667,0.000) -- (2.000,0.000) -- (2.333,1.270) -- (2.667,3.041) -- (3.000,3.560) -- (3.333,3.924) -- (3.667,4.221) -- (4.000,3.778) -- (4.333,4.149) -- (4.667,4.150) -- (5.000,4.485) -- (5.333,4.404) -- (5.667,4.555) -- (6.000,4.706) -- (6.333,4.746) -- (6.667,4.831) -- (7.000,4.698) -- (7.333,4.511) -- (7.667,4.738) -- (8.000,5.047) -- (8.333,5.078) -- (8.667,5.109) -- (9.000,5.262) -- (9.333,5.127) -- (9.667,5.363) -- (10.000,5.425);
\draw[thick,blue] (0.333,6.620)  -- (0.667,6.770) -- (1.000,6.656) -- (1.333,6.771) -- (1.667,6.711) -- (2.000,6.860) -- (2.333,6.955) -- (2.667,6.863) -- (3.000,6.940) -- (3.333,6.960) -- (3.667,6.872) -- (4.000,6.828) -- (4.333,6.890) -- (4.667,6.950) -- (5.000,6.968) -- (5.333,6.951) -- (5.667,6.872) -- (6.000,6.943) -- (6.333,6.945) -- (6.667,6.929) -- (7.000,7.007) -- (7.333,6.939) -- (7.667,6.972) -- (8.000,6.863) -- (8.333,6.989) -- (8.667,7.029) -- (9.000,7.070) -- (9.333,7.061) -- (9.667,7.023) -- (10.000,7.059);
\draw[thick,black] (0.333,7.127)  -- (0.667,7.454) -- (1.000,7.089) -- (1.333,7.116) -- (1.667,7.535) -- (2.000,7.180) -- (2.333,7.519) -- (2.667,7.397) -- (3.000,7.572) -- (3.333,7.529) -- (3.667,7.689) -- (4.000,7.634) -- (4.333,7.447) -- (4.667,7.592) -- (5.000,7.581) -- (5.333,7.641) -- (5.667,7.715) -- (6.000,7.549) -- (6.333,8.001) -- (6.667,8.192) -- (7.000,8.163) -- (7.333,7.905) -- (7.667,8.075) -- (8.000,8.262) -- (8.333,8.367) -- (8.667,8.192) -- (9.000,8.242) -- (9.333,8.390) -- (9.667,8.346) -- (10.000,8.346);
\end{tikzpicture}
\begin{tikzpicture}[scale=0.5]
\node at (5,-2) {Essay \#2};
\draw[thick,->] (0,0) -- (10,0);
\draw[thick,->] (0,0) -- (0,10);
\draw[black!40] (0,0) grid (10,10);
\node at (0,-1) {0};
\node at (-1,0) {0.3};
\node at (2,-1) {3k};
\node at (-1,2) {0.4};
\node at (4,-1) {6k};
\node at (-1,4) {0.5};
\node at (6,-1) {9k};
\node at (-1,6) {0.6};
\node at (8,-1) {12k};
\node at (-1,8) {0.7};
\node at (10,-1) {15k};
\node at (-1,10) {0.8};
\draw[thick, blue] (0.333,2.648)  -- (0.667,3.043) -- (1.000,3.152) -- (1.333,3.406) -- (1.667,3.654) -- (2.000,3.561) -- (2.333,3.567) -- (2.667,3.463) -- (3.000,3.587) -- (3.333,3.558) -- (3.667,3.484) -- (4.000,3.533) -- (4.333,3.414) -- (4.667,3.476) -- (5.000,3.399) -- (5.333,3.414) -- (5.667,3.407) -- (6.000,3.412) -- (6.333,3.423) -- (6.667,3.387) -- (7.000,3.352) -- (7.333,3.367) -- (7.667,3.426) -- (8.000,3.412) -- (8.333,3.462) -- (8.667,3.523) -- (9.000,3.494) -- (9.333,3.444) -- (9.667,3.479) -- (10.000,3.549);
\draw[thick, red] (0.333,0.000)  -- (0.667,0.000) -- (1.000,0.000) -- (1.333,0.000) -- (1.667,0.000) -- (2.000,0.000) -- (2.333,0.000) -- (2.667,0.000) -- (3.000,0.076) -- (3.333,0.042) -- (3.667,0.621) -- (4.000,0.624) -- (4.333,0.438) -- (4.667,0.430) -- (5.000,0.996) -- (5.333,0.568) -- (5.667,1.381) -- (6.000,1.329) -- (6.333,1.705) -- (6.667,1.675) -- (7.000,1.741) -- (7.333,1.345) -- (7.667,2.411) -- (8.000,2.317) -- (8.333,2.605) -- (8.667,2.842) -- (9.000,2.254) -- (9.333,3.330) -- (9.667,3.391) -- (10.000,3.561);
\draw[thick, black] (0.333,3.941)  -- (0.667,4.773) -- (1.000,5.745) -- (1.333,5.525) -- (1.667,5.541) -- (2.000,5.880) -- (2.333,5.734) -- (2.667,6.059) -- (3.000,6.149) -- (3.333,5.854) -- (3.667,5.543) -- (4.000,6.161) -- (4.333,6.265) -- (4.667,6.234) -- (5.000,6.403) -- (5.333,6.264) -- (5.667,6.147) -- (6.000,6.411) -- (6.333,6.014) -- (6.667,6.397) -- (7.000,6.599) -- (7.333,6.387) -- (7.667,6.541) -- (8.000,6.707) -- (8.333,6.254) -- (8.667,6.656) -- (9.000,6.435) -- (9.333,6.285) -- (9.667,6.508) -- (10.000,6.5);
\end{tikzpicture}

\begin{tikzpicture}
\draw[thick, blue] (0,0) --(.5,0);
\node at (1.2,0) {BOW};
\end{tikzpicture}\hspace{1cm}
\begin{tikzpicture}
\draw[thick, red] (0,0) --(.5,0);
\node at (1.2,0) {LSTM};
\end{tikzpicture}\hspace{1cm}
\begin{tikzpicture}
\draw[thick, black] (0,0) --(.5,0);
\node at (1.2,0) {BERT};
\end{tikzpicture}
\caption{The performance on the dimension of conventions. On the left we have the essay item \# 1, on the right we have essay item \#2. We have set a lower bound of 0.3 and an upper bound of 0.7 to better display the differences in the models.}\label{fig:val_conventions}
\end{figure}
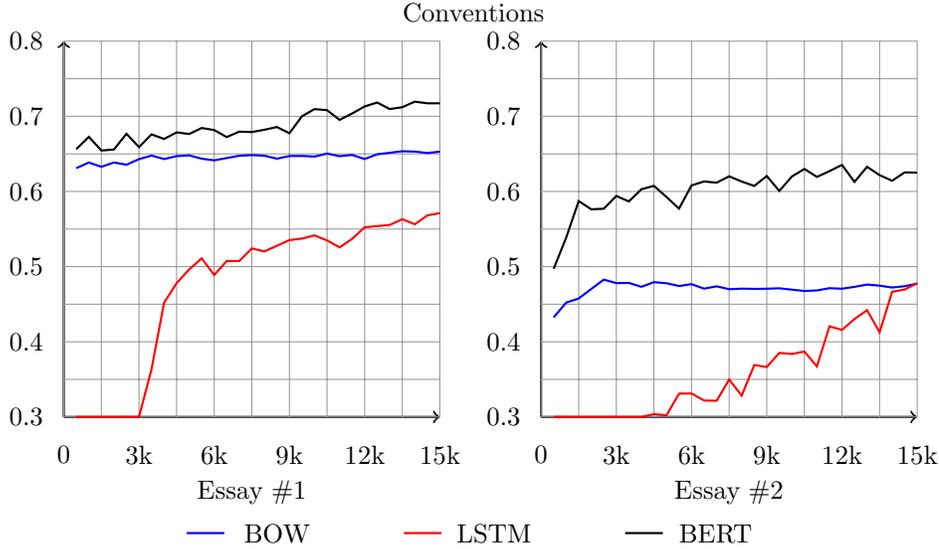

When we consider the performance on the trait of conventions, we see that BERT holds a considerable advantage across the board. Given we used only 16 textual features, the linear layer determining the classification in the BERT architecture is a function of 768 inputs that depend on the entire input space. While dissecting the features of a BERT model is difficult, we know that conventions could be interpreted as a measure of how discrepant the target text from the grammatically correct and impeccably spelled texts that BERT was exposed to in pretraining. It is clear that the features learned in this pretraining process seems to capture more features required to model conventions than the preprogrammed features of the BOW model. 

It is also clear that the LSTM models are at a distinct disadvantage. Since the fast-text embedding used was lower-case, it is clear that correct capitalization cannot be determined by the LSTM model. Furthermore, these models are trying to ascertain the rules of grammar from a small collection of texts. The ability to discern the correct words from incorrectly spelled words works against the ability of the LSTM to score conventions accurately. The fact that the LSTM falls short of the BOW and BERT models is no surprise in this context.

It should be noted that the spread of the data, as measured by the SMD, does not improve significantly with the amount of data used in any of our methods.

\begin{table}[!ht]
    \centering
    \begin{tabular}{|c|c|c | c c c |  c c c |} \hline
    &&& \multicolumn{3}{c|}{Single-Scores} &  \multicolumn{3}{c|}{Double-Scores}\\
        Prompt & Engine & Trait & QWK & SMD & Acc & QWK & SMD & Acc  \\ \hline \hline
  Essay \#1 & BOW & Elaboration & 0.810 & 0.047 & 73.5\% & 0.806 & 0.000 & 71.4\%\%\\
         & & Organization & 0.744 & 0.003 & 69.7\% & 0.774 & 0.028 & 71.4\%\\
         & & Conventions & 0.538 & 0.527 & 67.6\% & 0.733 & 0.022 & 76.8\%\\ \hline
          & LSTM & Elaboration & 0.804 & 0.140 & 68.5\% & 0.480 & 0.249 & 55.9\% \\
         & & Organization & 0.733 & 0.045 & 64.9\% & 0.472 & 0.323 & 60.3\% \\
         & & Conventions & 0.549 & 0.445 & 68.9\% & 0.468 & 0.463 & 65.1\%\\ \hline
          & BERT & Elaboration & 0.821 & 0.154 & 74.1\% & 0.806 & 0.042 & 73.7\%\\
         & & Organization & 0.741 & 0.191 & 69.3\%& 0.789 & 0.008 & 74.1\% \\
         & & Conventions & 0.573 & 0.530 & 69.9\% & 0.807 & 0.081 & 83.7\% \\ \hline \hline
  Essay \#2 & BOW & Elaboration & 0.689 & 0.218 & 63.4\% & 0.766 & 0.026 & 70.5\% \\
         & & Organization & 0.653 & 0.378 & 59.7\% & 0.821 & 0.005 & 75.5\% \\
         & & Conventions & 0.366 & 0.673 & 67.8\% & 0.727 & 0.110 & 79.4\% \\ \hline
          & LSTM & Elaboration & 0.793 & 0.182 & 67.6\% & 0.309 & 0.187 & 58.8\%\\
         & & Organization & 0.735 & 0.284 & 60.9\% & 0.269 & 0.035 & 54.5\% \\
         & & Conventions & 0.553 & 0.418 & 71.3\% & 0.338 & 0.652 & 67.4\%\\ \hline
          & BERT & Elaboration & 0.774 & 0.140 & 70.9\% & 0.791 & 0.056 & 72.1\%\\
         & & Organization & 0.731 & 0.289 & 64.0\% & 0.791 & 0.098 & 70.7\%\\
         & & Conventions & 0.557 & 0.529 & 71.7\% & 0.798 & 0.012 & 82.3\% \\ \hline
    \end{tabular}
    \caption{The statistics of the engines on the double scored validation data.}
    \label{tab:bigvsmall}
\end{table}

If we consider the results of the engines trained on a large set single-scored data, the results are poor when compared on the double-scored data for BOW and BERT and better for LSTM. If we are to believe that the double-scored data represents the interpretation of the rubric with the highest fidelity, then the SMD alone for conventions for each engine would disqualify the engines trained on this data for operational use. 

The data seems to indicate a mismatch between the labels of the single-scored data and the double-scored data. We need to bear in mind that the administrative conditions for the creation of these two datasets vary as the single-scored data was a corpus of responses that predated the use of AES for these two prompts. In any case, the labels are seen to have two distinct natures, hence, we expect some variation in the way in which the rubrics were interpreted. One score being more leniently in one administrative setting than the other causing a difference in the spread of scores, which can be seen in Table \ref{tab:spread}.

\begin{table}[]
    \centering
    \begin{tabular}{|c|c || c| c | c| c || c| c|}\hline
    && \multicolumn{4}{c||}{Training set}& \multicolumn{2}{c|}{Validation}\\ \hline
    && \multicolumn{2}{c|}{SS} & \multicolumn{2}{c||}{DS Train}& \multicolumn{2}{c|}{DS Val}\\ \hline 
        Essay & Trait  & mean & std & mean & std & mean & std \\ \hline \hline
        \# 1 & Elaboration & 1.611 & 0.925 &1.601 &0.878 &1.603 & 0.885 \\
         & Organization & 1.677 &0.852 &1.676 &0.829 & 1.689 & 0.844 \\
          & Conventions & 1.609 &0.652 &1.426&0.711& 1.468 & 0.690 \\ \hline
    \# 2 & Elaboration  & 1.740 & 0.970 & 1.736 & 0.857 & 1.736 & 0.861 \\
         & Organization & 1.996 & 0.969 & 1.900 & 0.864 & 1.898 & 0.849\\
          & Conventions & 1.689 & 0.566 & 1.520 & 0.676 & 1.541 & 0.664\\ \hline
    \end{tabular}
    \caption{The mean and standard deviations of the training sets and validation set in the results of Table \ref{tab:bigvsmall}}
    \label{tab:spread}
\end{table}

There is just one more possibility we wish to explore, that we use the single-scored data to define an initial state for the training of models to be trained on the double-scored data. The last part of the second experiment involves using the single-scored data for pretraining. This approach is similar to the work done on classifying tweets \cite{tweets}. The main idea is that the weaker data is used to define an appropriate set of features that were not abundant in the smaller dataset, from which the linear layer may use to more accurately classify the smaller dataset. This differs from the pretraining the underlying model as a language model since we do not use the LSTM to predict missing or future words, as done in \cite{pretrainLSTM}. Furthermore, the pretraining set bears a more accurate resemblance to the target dataset in this case. The results of this process are outlined in Table \ref{tab:transfer}.

\begin{table}[]
    \centering
    \begin{tabular}{|c| c| c ||c | c | c|} \hline 
    Prompt & Engine & trait & QWK & SMD & Acc \\ \hline \hline 
    \# 1 & BERT & Elaboration & 0.811 & 0.012 & 72.2\% \\ 
         && Organization &0.798 & 0.018 & 73.3\%\\
         && Conventions  &0.767 & 0.044 & 79.1\% \\ \hline
     & LSTM & Elaboration& 0.822 & 0.051 & 74.1\% \\ 
         && Organization  & 0.778 & 0.039 & 72.2\% \\
         && Conventions   & 0.549 & 0.445 & 68.9\% \\ \hline \hline
    \# 2 & BERT & Elaboration & 0.811 & 0.005 & 73.0\%\\ 
         && Organization &0.829 & 0.052 & 73.4\% \\
         && Conventions  &0.773 & 0.006 & 80.5\%  \\ \hline
     & LSTM & Elaboration & 0.773 & 0.044 & 71.9\%\\ 
         && Organization & 0.816 & 0.075 & 74.2\% \\
         && Conventions  & 0.668 & 0.1625 & 75.5\%\\ \hline
    \end{tabular}
    \caption{The results of using the models trained on the single scored data as the initial state for models trained on the double-scored data.}
    \label{tab:transfer}
\end{table}

It is interesting to see that BERT seems to see increases in the performance characteristics for elaboration and organization when subjected to this training and decreases in conventions. We speculate that this may be a symptom of the system forgetting much of the pretraining in the process of training on the single-scored data. Elaboration and organization, on the other hand, are based on features that are more specific to the prompt. What is also interesting is how comparable the LSTM based engine performs given the engine itself (not including the embedding) possesses approximately 3.5 million parameters, while the BERT model possesses 110 million parameters (23 million for the embedding).

\section{Discussion}\label{sec:discussion}

One of the aspects of this study we did not delve into is the ability to tune the parameters that define neural networks. Tuning parameters such as learning rate, batch-size, dropout, and recurrent dropout can lead to significant improvements in text-classification results \cite{hyperparameter}. We expect that hyperparameter tuning should improve the performance of BERT and the LSTM-model significantly on all traits and even exceed human performance. Typically, we see in the order of 5-10\% improvement due to hyperparamater tuning. When tuning these parameters, we often rely on simple grid search methods as well as Baysean approaches \cite{sigopt}. The only parameter we can tune in the case of the BOW-model chosen is the LSA-dimension. 

The other aspect not touched is the pretraining of an LSTM, the difficulty in doing so has been reported in \cite{pretrainLSTM}. We believe that similar results to the BERT and BOW models are possible by pretraining the LSTM as a part of an autoencoder network. It is worth noting that the current state-of-the-art results on the Kaggle dataset (see \cite{Kaggle}) is achieved with an attention-based ensemble of convolutional and LSTM units \cite{AttAES}, hence, we can assume the addition of an attention mechanism should improve the results significantly. It may also be able to do a kind of pretraining for the LSA by using the LSA features defined by a larger dataset which is then used as the features for a classifier for a smaller dataset. 

An interesting aspect of the conventions trait is that it is based on universal features associated with grammar, spelling, and prose. From an educational standpoint, how these features are assessed depend upon the intended grade this prompt is given; however, it is the same set of features that are taken into consideration for each grade. This means it might be possible to pool the data from multiple prompts, meaning that it may be possible to provide an LSTM with sufficient data to perform comparably on the conventions trait, especially on an embedding that is case-sensitive. 

\section{Conclusion}

When we initiated this line of research, the consensus seemed to be that single-scored data would provide little to no value in developing an effective AES engine. This seems to be true for traditional AES models built on an ensemble of hand-crafted and LSA features, however, the possibility of transfer learning makes single-scored data useful. This study has illuminated a few things; LSTMs seem to be able to perform comparably to transformer-based and BOW-based models in elaboration and organization with enough data, that conventions seems to benefit more from the initial features defined in the pretrained model.

\end{document}